\documentclass[lettersize,journal]{IEEEtran}
\usepackage{amsmath,amsfonts}
\usepackage{algorithmic}
\usepackage{array}
\usepackage[caption=false,font=normalsize,labelfont=sf,textfont=sf]{subfig}
\usepackage{textcomp}
\usepackage{stfloats}
\usepackage{url}
\usepackage{verbatim}
\usepackage{graphicx}
\def\BibTeX{{\rm B\kern-.05em{\sc i\kern-.025em b}\kern-.08em
    T\kern-.1667em\lower.7ex\hbox{E}\kern-.125emX}}
\usepackage{balance}

\usepackage[colorlinks,urlcolor=blue,linkcolor=blue,citecolor=blue]{hyperref}

\usepackage{color,array}
\usepackage{amsmath}
\usepackage{amsthm}
\usepackage{graphicx}
\usepackage{booktabs}
\usepackage{multirow}
\usepackage{subcaption}
\usepackage{orcidlink}
\usepackage{algorithm}
\usepackage{algorithmic}
\usepackage{balance}
\usepackage{amssymb}
\usepackage[switch]{lineno}
\usepackage{threeparttable}
\usepackage{tikz}
\usepackage{listings}

\usepackage[normalem]{ulem}
\useunder{\uline}{\ul}{}

\usepackage{xcolor}

\definecolor{codegray}{gray}{0.95}
\definecolor{darkblue}{rgb}{0.1,0.1,0.6}
\definecolor{darkgreen}{rgb}{0.0,0.5,0.0}
\definecolor{darkred}{rgb}{0.6,0.1,0.1}

\lstdefinestyle{mystyle}{
  language=Python,
  backgroundcolor=\color{codegray},
  basicstyle=\footnotesize\ttfamily,
  keywordstyle=\color{darkblue}\bfseries,
  stringstyle=\color{darkred},
  commentstyle=\color{darkgreen}\itshape,
  identifierstyle=\color{black},
  showstringspaces=false,
  numberstyle=\tiny\color{gray},
  numbers=left,
  numbersep=5pt,
  frame=single,
  breaklines=true,
  captionpos=b,
  tabsize=2,
}

\newcommand{\pie}[1]{%
\begin{tikzpicture}
 \draw (0ex,0ex) circle (1ex);
 \fill (0ex,-1ex) arc (-90:(#1-90):1ex) -- (0ex,-1ex) -- cycle;
\end{tikzpicture}%
}

\begin{document}

\title{ABE: A Unified Framework for Robust and Faithful Attribution-Based Explainability}

% \author{Anonymous Authors}

\author{
Zhiyu Zhu\textsuperscript{1},
Jiayu Zhang\textsuperscript{2},
Zhibo Jin\textsuperscript{1},
Fang Chen\textsuperscript{1},
Jianlong Zhou\textsuperscript{1}
\\
\textsuperscript{1}University of Technology Sydney, Sydney, NSW, Australia\\
Emails: \{zhiyu.zhu,zhibo.jin\}@student.uts.edu.au, \{fang.chen,jianlong.zhou\}@uts.edu.au\\
\textsuperscript{2}Changshu Institute of Technology, Changshu, China\\
Email: zjy@cslg.edu.cn
}

\markboth{}
{}

\maketitle

\begin{abstract}
Attribution algorithms are essential for enhancing the interpretability and trustworthiness of deep learning models by identifying key features driving model decisions. Existing frameworks, such as InterpretDL and OmniXAI, integrate multiple attribution methods but suffer from scalability limitations, high coupling, theoretical constraints, and lack of user-friendly implementations, hindering neural network transparency and interoperability. To address these challenges, we propose Attribution-Based Explainability (ABE), a unified framework that formalizes Fundamental Attribution Methods and integrates state-of-the-art attribution algorithms while ensuring compliance with attribution axioms. ABE enables researchers to develop novel attribution techniques and enhances interpretability through four customizable modules—Robustness, Interpretability, Validation, and Data \& Model. This framework provides a scalable, extensible foundation for advancing attribution-based explainability and fostering transparent AI systems. Our code is available at: \hyperlink{https://github.com/LMBTough/ABE-XAI}{https://github.com/LMBTough/ABE-XAI}.

\end{abstract}


\section{Introduction}

With the widespread application of deep learning models, neural networks have demonstrated remarkable performance across various tasks, including image classification, natural language processing, and speech recognition. However, despite the significant progress made in terms of accuracy and efficiency, the "black-box" nature of these models has made the interpretability of model decisions particularly crucial. The lack of transparency not only hinders the deployment of models in high-risk scenarios, such as medical diagnosis and financial decision-making, but also limits researchers' ability to understand and improve the internal workings of these models. Therefore, enhancing the interpretability of neural networks is vital for building trustworthy artificial intelligence systems~\cite{zhou2021evaluating}.

Research on AI interpretability categorizes methods into those that satisfy attribution axioms~\cite{sundararajan2017axiomatic} and those that do not (see Section~\ref{apx:related}). Attribution-based methods, such as Integrated Gradients (IG)~\cite{sundararajan2017axiomatic} and Boundary-based Integrated Gradients (BIG)~\cite{wang2021robust}, offer more transparent and stable explanations than traditional techniques like Grad-CAM~\cite{selvaraju2017grad}, enabling fine-grained feature-level interpretation. However, they are constrained by linear exploration paths and often overlook model robustness. Recent advances integrate adversarial attack-based attributions~\cite{pan2021explaining,zhu2024mfaba,zhuattexplore}, linking interpretability with robustness—a key metric for assessing a model’s resilience to input perturbations. By analyzing decision shifts under adversarial disturbances, these methods identify critical features without relying on reference inputs (baseline inputs such as a black image or a zero vector), thereby enhancing explanation consistency and robustness. Since adversarial attacks cause minimal perturbations that significantly impact decisions, they highlight essential features for model reasoning. Incorporating robustness into attribution analysis improves explanation reliability and provides a theoretical foundation for robustness testing, advancing both interpretability and model evaluation.

However, existing attribution frameworks~\cite{NIPS2017_7062,yang2022omnixai,li2022interpretdl} exhibit several significant limitations in terms of usability, integration, and task support. Most current frameworks are designed for specific task types, such as classification, and lack the flexibility to accommodate multimodal tasks. Moreover, these frameworks often struggle to effectively integrate new attribution methods that adhere to established attribution axioms, thereby hindering the consistency and reliability of attribution results. The absence of standardized evaluation metrics further complicates the comparison and assessment of different attribution techniques, contributing to the complexity of the attribution evaluation system. Moreover, end-users often face the challenge of selecting an appropriate explanation method from dozens of alternatives without clear guidance. The lack of standardized metrics and robustness-aware evaluation further complicates this process, leaving users without a principled way to identify the most trustworthy attribution for their specific task.

To address these challenges, we propose \textbf{ABE}, a unified framework for \textbf{A}ttribution-\textbf{B}ased \textbf{E}xplainability that integrates both traditional and state-of-the-art attribution methods under a principled evaluation and robustness perspective. At the core of ABE is a set of \textbf{Fundamental Attribution Methods} that unify gradient-based attribution techniques into a flexible and theoretically grounded formulation, enabling consistent integration of diverse update strategies while preserving attribution axioms. A central feature of ABE is its \textbf{robustness module}, which incorporates a series of update methods that enhance explanation stability by simulating minimal perturbations and measuring their influence on model decisions. These update methods are seamlessly integrated into a modular architecture that enables interoperability between attribution algorithms, robustness testing procedures, and evaluation pipelines. The framework supports a wide variety of model types and task modalities, including both unimodal and multimodal scenarios. Its modular design allows users to customize attribution workflows, select from a diverse set of metrics, and integrate new attribution or evaluation methods without modifying the core structure. This enables users to perform flexible and reliable interpretability analysis while maintaining theoretical consistency with attribution axioms. Moreover, ABE includes essential utility modules for data loading, model interfacing, attribution generation, and result validation, ensuring usability and reproducibility. By supporting extensible components, ABE provides a comprehensive solution for advancing interpretable and robust machine learning research.

\begin{itemize}

\item[1.] We design a class of \textbf{Fundamental Attribution Methods}, which abstract attribution axioms into a general gradient-based formulation. This design allows flexible incorporation of custom update strategies while preserving theoretical guarantees such as completeness and sensitivity.

\item[2.] We introduce a \textbf{Robustness Module} that augments attribution computation with perturbation-aware update methods, enabling explanation stability analysis under adversarial variations. This supports the automatic selection of the most robust and faithful explanations based on quantitative metrics such as insertion and deletion scores.

\item[3.] We propose \textbf{ABE}, a unified and extensible framework for attribution-based explainability, which integrates both classical and adversarial attribution methods, enabling principled robustness evaluation and method selection under a common interface.

\item[4.] We implement the framework as a modular PyTorch-based toolkit supporting multimodal tasks. It provides reusable components for model interfacing, attribution computation, evaluation, and visualization, facilitating practical use, extensibility, and reproducibility in interpretability research.

\end{itemize}

\section{Related Work} \label{apx:related}
\subsection{Traditional Interpretability Methods}
Traditional interpretability methods primarily aim to address the "black-box" problem of deep learning models by providing intuitive explanations of their decision-making processes, thus enhancing model transparency. Saliency Maps~\cite{simonyan2013deep}, an early gradient-based method, identify important regions by calculating the gradient of the output with respect to input features. However, their results are sensitive to noise and gradient vanishing issues, leading to unstable attributions.

SmoothGrad (SG)~\cite{smilkov2017smoothgrad} improves saliency maps by applying multiple random perturbations to input samples and averaging the attribution results, which reduces the impact of noise and enhances stability. However, SG’s attribution performance is dependent on the perturbation strategy and is computationally expensive. Layer-wise Relevance Propagation (LRP)~\cite{binder2016layer} and DeepLIFT~\cite{shrikumar2017learning} assign relevance scores to model outputs via backpropagation, reducing noise interference and providing more interpretable results in deep networks, though their attribution results are model-specific.

Gradient-weighted Class Activation Mapping (Grad-CAM)~\cite{selvaraju2017grad} leverages convolutional neural network feature maps to generate more intuitive heatmaps, highlighting key regions the model focuses on. However, it relies on activation information from intermediate layers and has limited performance in non-convolutional networks. The Shapley value method~\cite{lundberg2017unified}, based on game theory, offers a theoretically fair framework for distributing feature contributions. Despite its strong consistency, it suffers from high computational complexity. Local Interpretable Model-agnostic Explanations (LIME)~\cite{ribeiro2016should} constructs linear surrogate models by perturbing local samples, offering flexibility and applicability to various data types, without relying on model structure.

While these traditional methods have made significant strides in explaining model decisions, they still have limitations, such as overlooking interactions between features, which can miss key feature contributions. Additionally, the attribution results are prone to model implementation details and input perturbations, resulting in instability and inconsistency.

\begin{table*}[t]
\caption{Comparison of Interpretability Frameworks}
\label{tab:explainability_frameworks}
\resizebox{\textwidth}{!}{%
\begin{tabular}{@{}l|l|l|l|l|l@{}}
\toprule
Library       & Algorithms                                    & Framework    & Model Type                                                              & Benchmark                                              & Metrics                                                                   \\ \midrule
SHAP~\cite{NIPS2017_7062}          & \parbox{4cm}{Shapley Values}                  & Python       & \parbox{4cm}{Tabular, Text, Image}                                      & \parbox{4cm}{Feature Attribution, Feature Importance} & \parbox{4cm}{Shapley Value Score, Mean Absolute Error (MAE)}            \\ \midrule
Captum~\cite{kokhlikyan2020captum}        & \parbox{4cm}{Integrated Gradients, DeepLIFT, LRP, Grad-CAM} & PyTorch      & \parbox{4cm}{Classification, Regression}                                & \parbox{4cm}{Sensitivity Analysis, Saliency Maps}      & \parbox{4cm}{Attribution Accuracy, Insertion/Deletion Scores}           \\ \midrule
Dattri~\cite{deng2024dattri}        & \parbox{4cm}{IF, TracIn, RPS, TRAK}           & PyTorch      & \parbox{4cm}{Tabular, Image, Text, Music}                               & \parbox{4cm}{Feature Interaction, Noisy Label Detection} & \parbox{4cm}{Label Noise Detection Rate, Feature Interaction Strength}  \\ \midrule
Interpret~\cite{nori2019interpretml}     & \parbox{4cm}{SHAP, Lime, TreeSHAP}            & Python       & \parbox{4cm}{Tabular, Text, Image}                                      & \parbox{4cm}{Global \& Local Explanation}              & \parbox{4cm}{Model Fidelity, Consistency, Explanation Stability}        \\ \midrule
InterpretDL~\cite{JMLR:v23:21-0738}   & \parbox{4cm}{Integrated Gradients, Grad-CAM, SmoothGrad} & PaddlePaddle & \parbox{4cm}{Image}                                                   & \parbox{4cm}{Image Attribution}                        & \parbox{4cm}{Saliency Map Quality, Robustness Score}                    \\ \midrule
OmniXAI~\cite{wenzhuo2022-omnixai}       & \parbox{4cm}{Multiple Attribution Methods}    & Python       & \parbox{4cm}{Multimodal}                                              & \parbox{4cm}{Classification, Regression, Time Series}  & \parbox{4cm}{Model Accuracy, Attribution Consistency}                   \\ \midrule
ABE (Ours) & \parbox{4cm}{Foundational Attribution Method, Robustness Module, 17 interpretability methods, Highly customizable} & PyTorch      & \parbox{4cm}{Image Classification, Text Classification, Multimodal, Object Detection} & \parbox{4cm}{Attribution, Robustness Evaluation, Multimodal Support} & \parbox{4cm}{Insertion \& Deletion Scores, Confidence Increase \& Drop, INFD Score, Time Consumption} \\ \bottomrule
\end{tabular}%
}
\end{table*}

\subsection{Attribution Axiom-Based Interpretability Methods}
To overcome the limitations of traditional interpretability methods in terms of stability and consistency, recent research has proposed axiom-based approaches to improve the reliability of model decision explanations. The Integrated Gradients (IG)~\cite{sundararajan2017axiomatic} method introduced two key attribution axioms: Sensitivity and Implementation Invariance. The Sensitivity axiom requires that when a feature differs from its baseline and causes a change in the prediction, the attribution method must assign a non-zero attribution value to that feature, reflecting its importance. The Implementation Invariance axiom states that if two networks produce identical outputs for all inputs, the attribution results on both networks should also be identical, regardless of their internal implementations. These axioms significantly enhance the stability and consistency of attribution results, making the explanations more reliable. We have designed a fundamental attribution method that, when applied, enables other attribution methods to satisfy the attribution axioms as well. Details can be found in Section~\ref{sec:fundamental}.

Although IG satisfies these axioms, it has certain limitations. IG uses a fixed baseline (e.g., a zero-vector or black image) and calculates attributions along a straight-line path. This fixed-path attribution is susceptible to baseline choices, leading to instability in attribution results. To address this, the Boundary-based Integrated Gradients (BIG)~\cite{wang2021robust} method dynamically selects baselines and integrates multiple paths to improve attribution robustness. Unlike IG, BIG does not rely on a single baseline but flexibly selects one based on input data features and combines attributions from multiple paths, offering more stable and consistent explanations across different input types and model architectures.

Furthermore, the Expected Gradients (EG)~\cite{erion2021improving} method improves IG by sampling multiple baselines from the input feature distribution and computing the expected attribution values for these baselines. This reduces the uncertainty introduced by selecting a single baseline, improving attribution robustness and generalization. To reduce the computational cost of path integration, Fast Integrated Gradients (FIG)~\cite{hesse2021fast} proposes a fast approximation method, significantly reducing the number of integration points, thus enhancing computational efficiency while maintaining attribution accuracy. Guided Integrated Gradients (GIG)~\cite{kapishnikov2021guided} combines IG and Guided Backpropagation, suppressing the influence of negative gradients and emphasizing the contribution of positive gradients to model predictions, generating clearer and more intuitive attribution maps. These methods mitigate the limitations of IG in various scenarios, yet they remain less sensitive to input perturbations and may not capture feature changes near decision boundaries. Additionally, most of these methods focus on single-model explanations, neglecting the challenges of cross-model consistency. Therefore, solely relying on axiom-based methods may not fully reveal the relationship between model robustness and interpretability, prompting researchers to explore integrating attribution with robustness testing.

\subsection{Robustness and Attribution Methods}
The Adversarial Gradient Integration (AGI)~\cite{pan2021explaining} method was the first to combine model robustness with attribution. AGI constructs the steepest ascent path between adversarial samples and the original input, revealing decision changes in response to adversarial perturbations. This path selection captures sensitive features near the model’s decision boundary and better reflects model robustness. The core idea of AGI is to use gradients from adversarial samples to explore nonlinear paths for attribution, bypassing the dependency on reference inputs found in traditional methods. This makes the attribution process more consistent and systematic, enhancing the robustness of the explanation results. However, AGI still has some shortcomings. First, its path selection focuses on the decision boundary of a single model, lacking cross-model consistency. Second, AGI’s adversarial path exploration is limited to targeted attacks, which restricts its exploration range and prevents it from capturing feature variations under different input perturbations.

MFABA~\cite{zhu2024mfaba} combines adversarial perturbations and boundary attribution, enhancing the robustness of traditional methods. By introducing more efficient boundary point calculations, MFABA improves attribution accuracy and stability when handling adversarial samples. However, like AGI, MFABA relies on the basic Boundary Iterative Method (BIM)~\cite{kurakin2018adversarial} for path exploration, which limits its exploration range and diversity. Although MFABA improves robustness in some cases, its limited path exploration may not fully capture feature changes under varying input perturbations.

AttEXplore~\cite{zhuattexplore} addresses the shortcomings of AGI and MFABA and further elucidates the connection between black-box attacks and interpretability. AttEXplore explores multiple model parameter spaces, generating consistent paths across models, thus enhancing the transferability and stability of explanations. Unlike AGI and MFABA, AttEXplore’s path exploration is more diverse, covering a broader decision boundary and ensuring consistent attribution explanations across different models and scenarios. Moreover, AttEXplore incorporates frequency-domain information, enabling the attribution process to better resist input noise and perturbations, thus providing more robust and comprehensive explanation results.

\subsection{Interpretability Frameworks}
In recent years, with the increasing complexity of deep learning models, various interpretability frameworks have been proposed to provide efficient and flexible interpretability tools for different models and tasks. These frameworks cover a range of methods, from traditional attribution techniques to interpretability methods for multimodal tasks, showcasing their strengths and limitations across different application scenarios. To provide a more intuitive comparison of these frameworks, Table \ref{tab:explainability_frameworks} summarizes key information, including supported algorithms, framework types, applicable model types, benchmark tasks, and evaluation metrics.

Our framework introduces significant improvements over the aforementioned frameworks, combining 17 attribution methods (including several of the latest approaches) with robustness modules. In addition to implementing Insertion Score, Deletion Score, Confidence Increase, Confidence Drop, and INFD Score for comprehensive model interpretability evaluation, our framework supports multimodal tasks, including image and text classification, as well as object detection tasks.

\section{Method}
\subsection{Problem Definition} \label{sec:pd}

\begin{figure*}
    \centering
    \includegraphics[width=.8\linewidth]{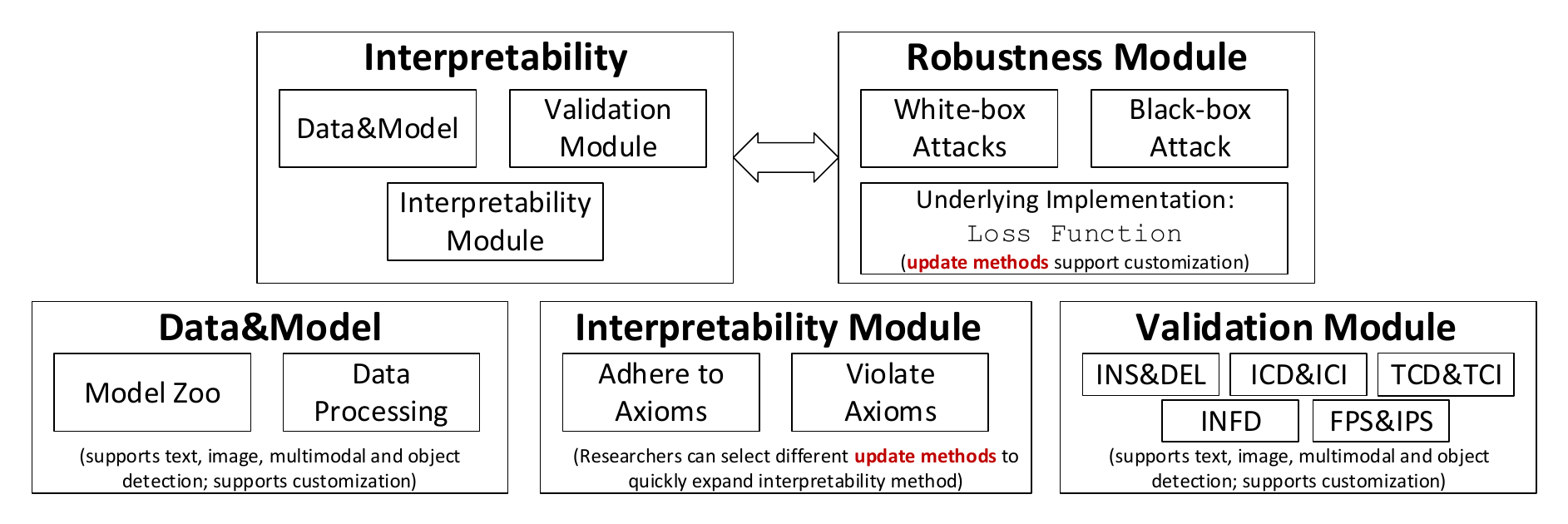}
    \caption{Flowchart of our Framework}
    \label{fig:flow}
\end{figure*}

In this section, we first formally define the neural network and its related interpretability concepts, including attribution methods, axiomatic constraints, and robustness. Based on this foundation, we introduce the definition of adversarial attacks to further illustrate the relationship between robustness and attribution.

\paragraph{Definition of Neural Network} A neural network can be defined as a function $f: \mathbb{R}^d \rightarrow \mathbb{R}^k$, where $d$ represents the dimension of the input features, and $k$ denotes the number of output classes. For a given input vector $x \in \mathbb{R}^d$, the neural network outputs a prediction vector $f(x) \in \mathbb{R}^k$, where each element $f(x)_i$ represents the confidence of the input belonging to class $i$. Here, $x_j$ represents the feature value of the $j$-th dimension of the input vector $x$.

\paragraph{Definition of Interpretability and Attribution} Interpretability aims to provide understandable explanations for the model's predictions. In attribution methods, given an input sample $x$ and the model's prediction $f(x)$, the goal of attribution methods is to compute the contribution of the input feature $x_j$ to the model output. The attribution value can be represented as $A_j(x)$, where $A_j(x)$ represents the contribution of feature $x_j$ to the output.

Attribution methods often require a reference input $x'$, and the attribution result is represented as the attribution vector $A(x, x')$, where $A_j(x, x')$ indicates the contribution of the $j$-th dimension of feature $x_j$ during the transformation process from $x'$ to $x$.

\paragraph{Definition of Attribution Axioms} To ensure the reliability of attribution results, attribution methods must satisfy the following two core axioms~\cite{sundararajan2017axiomatic}. Additionally, we also use the \textit{impossibility theorems}~\cite{bilodeau2024impossibility} in the analysis, particularly focusing on the \textbf{Linear and Complete} properties.

\textit{Sensitivity Axiom:} The Sensitivity Axiom requires that when there is a difference between the input feature $x_j$ and the reference input $x'_j$, and this difference affects the model's output, the attribution value $A_j(x, x')$ must be non-zero. This is formally defined as: 
\begin{equation} 
\text{If} \quad f(x) \neq f(x') \quad \text{and} \quad x_j \neq x'_j, \quad \text{then} \quad A_j(x, x') \neq 0.
\end{equation}

\textit{Implementation Invariance Axiom:} The Implementation Invariance Axiom requires that if two models $f_1$ and $f_2$ produce the same output for all inputs, i.e., $f_1(x) = f_2(x), \forall x \in \mathbb{R}^d$, then the attribution results of the attribution method should also be the same for both models, i.e., $A(x, x')_{f_1} = A(x, x')_{f_2}$. If only one model is considered, for simplicity, we will omit the subscripts $f_1$ and $f_2$ in the notation.

% \textit{Definition of impossibility theorems} \label{apx:impossible}

\textit{Definition of Complete:}
A feature attribution method $A$ is said to be \emph{complete} if, for a given input $x$ and reference input $x'$, the sum of the attributions for all features equals the difference in the model's output between $x$ and $x'$. Formally,
\begin{equation}
\sum_{j \in [p]} A_j(x, x') \;=\; f(x) \;-\; f(x')
\end{equation}
where $A_j(x, x')$ denotes the attribution of the $j$-th feature. The term $f(x) - f(x')$ indicates how much the model's output changes when moving from the reference input $x'$ to the target input $x$.

\textit{Definition of Linear:}
A feature attribution method $A$ is said to be \emph{linear} if, for any linear model $f(x) = \sum_{j \in [p]} f_j ( x_j)$, where the model $f$ is composed of a sum of $p$ functions, each depending on a single input feature, the attribution assigned to each feature $x_j$ is exactly $\theta_j \, x_j$.

\begin{equation}
A(x,x^\prime ,f)_j = A(x_j,x_j^\prime ,f_j)
\end{equation}

When the underlying model is linear, the attribution to each feature is directly proportional to that feature's contribution in the linear function.

\paragraph{Definition of Robustness} Robustness refers to the ability of a model to maintain stable predictions when faced with input perturbations. Formally, robustness can be defined as the model's ability to satisfy the following condition:

\begin{equation} 
\label{eq:rubust}
\forall x \in X, \forall x^{adv} \in B_{\epsilon }(x), \quad f(x^{adv}) = f(x)
\end{equation}

This definition requires that for any input $x$ in the input space $X$ and any adversarial example $x^{adv}$ within the $\epsilon$-ball $B_{\epsilon}(x) = \{ x' \mid x' \in [x - \epsilon, x + \epsilon] \}$, where $\epsilon$ denotes the maximum allowable perturbation, the model's output should remain unchanged. In other words, a robust model is expected to produce consistent predictions under small perturbations to the input.In practice, absolute robustness does not exist. Therefore, we evaluate a model's robustness by measuring the number of samples that satisfy Equation~\ref{eq:rubust}.

Adversarial attacks are a common method used to test the robustness of a model. An adversarial sample $x^{adv}$ is defined as the result of the following maximization in the case of an untargeted attack:

\begin{equation} 
\max_{x^{adv} \in B_{\epsilon }(x)} L_{y}(x^{adv})
\end{equation}

Where $L$ is the loss function, and in classification tasks, maximizing $\max_{x^{adv} \in B_{\epsilon}(x)} L_{y}(x^{adv})$ entails finding an adversarial sample $x^{adv}$ in the $\epsilon$-ball $B_{\epsilon}(x)$ that increases the model’s loss with respect to the true label $y$. By increasing $L_{y}(x^{adv})$, the attacker seeks to reduce the model’s confidence for the correct label, potentially causing a misclassification. This reveals the most “vulnerable” features the model relies on for accurate predictions. If $f(x^{adv}) \neq f(x)$, it signifies that the perturbation successfully alters the model’s output, highlighting critical features pivotal to the model’s decision boundary. \textbf{When two input samples are very close but yield different outputs, these subtle perturbations can be the key reason their classes differ.} Such adversarial samples also serve as a valuable tool for measuring robustness (by testing model performance on adversarial inputs) and can be employed as reference inputs $x'$ in interpretability methods to further analyze which features drive the model’s predictions.

\subsection{Module Design}

The figure~\ref{fig:flow} presents a framework comprising four key modules: Interpretability, Robustness, Data\&Model, and Validation. The Interpretability Module focuses on adhering to or violating attribution axioms, supporting different update methods for customized interpretability. The Robustness Module integrates white-box and black-box attacks, while the Validation Module evaluates metrics like Insertion Score (INS), Deletion Score (DEL), and others for model validation and customization.

\paragraph{Fundamental Attribution Methods} \label{sec:fundamental}
In designing the Fundamental Attribution Methods, we have summarized the current mainstream attribution methods that satisfy the attribution axioms~\cite{sundararajan2017axiomatic}. The core requirement of these methods is the exploration of the sample space and the accumulation of gradients during this exploration, ensuring that the attribution results accurately reflect the impact of input features on the model's output.

\begin{figure}[h]
    \centering
    \includegraphics[width=\linewidth]{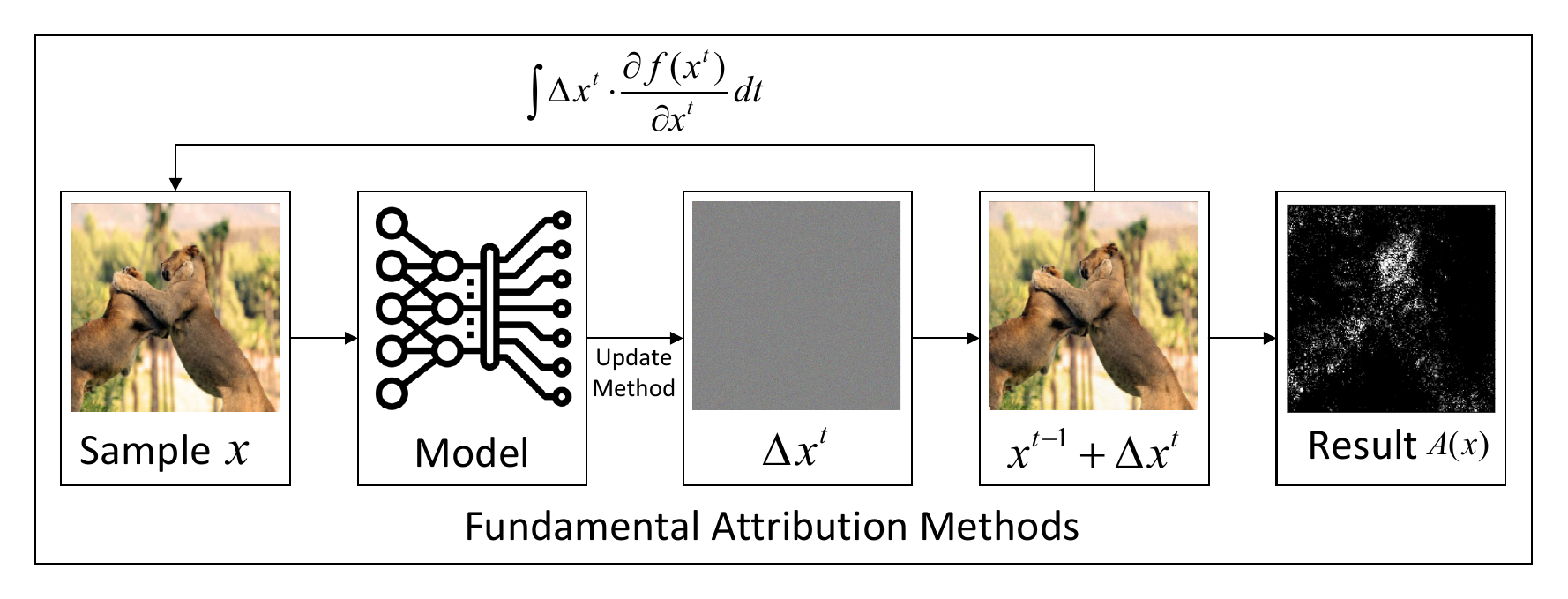}
    \caption{Flowchart of Fundamental Attribution Methods.}
    \label{fig:fundamental_flowchart}
\end{figure}

A key insight here is that we have abstracted the essential part of the attribution axioms into a unified theoretical framework, such that any user (by following this framework) can incorporate a wide variety of \emph{update} methods (the specific way of perturbing or moving from a baseline to the target). This guarantees that the resulting attribution method will still satisfy the required axioms. In other words, this framework provides theoretical assurance that third-party interpretability methods can integrate arbitrary update strategies yet maintain adherence to the fundamental attribution axioms.

The design of the fundamental attribution method follows:
\begin{equation}
\label{eq:loss}
\sum_{j=1}^{d} A_j(x, x') \;=\; \int \Delta x^t \cdot \frac{\partial f(x^t)}{\partial x^t} \, dt,
\end{equation}
where $x^t = x^t + \sum_{i=1}^{t} \Delta x^t$, and $\Delta x^t$ is obtained through the Update Method. In this formulation, each $\Delta x^t$ must lie in the local neighborhood of the current point $x^t$, ensuring that the Taylor expansion (or local gradient information) accurately captures the incremental changes to the model’s output. 

Equation~\ref{eq:loss} ensures that the cumulative attribution values equal the total effect of moving from the reference input $x'$ to the target input $x$. Importantly, this equation does not place any restriction on how $x$ is updated at each step; users can define custom update functions while ensuring that the resulting attribution method still satisfies the attribution axioms. This flexibility also guarantees that the method satisfies the \emph{Complete} property. Equation~\ref{eq:loss} corresponds to the fourth line of Algorithm~\ref{ag:funda}, where the attribution values are iteratively computed based on the model's output changes as the Update Method proceeds.

Figure~\ref{fig:fundamental_flowchart} illustrates the general framework of Fundamental Attribution Methods, which aim to approximate the integrated gradient of a model’s output with respect to the input. Given an input $x$, the algorithm iteratively updates the input vector $x^t$ using a pre-defined update method (e.g., linear or non-linear interpolation), and accumulates attribution $A^t$ over $T$ steps. At each iteration, the change in input $\Delta x^{t-1}$ is computed, and the corresponding gradient $\nabla f(x^{t-1})$ is used to update the attribution via $A^t = A^{t-1} + \Delta x^{t-1} \cdot \nabla f(x^{t-1})$, where $A^t$ denotes the attribution result at step $t$. The process can be viewed as a numerical approximation of the path integral $\int \Delta x^t \cdot \frac{\partial f(x^t)}{\partial x^t} dt$, and results in the final attribution map $A(x)$ after $T$ steps.

\begin{algorithm}[h]
\caption{Fundamental Attribution Methods}
\label{ag:funda}
\begin{algorithmic}[1]
\REQUIRE Number of attribution iterations $T$, target model f, input vector x
\ENSURE $A$
\STATE $A^0 = 0$
\FOR {$t = 1, 2, \dots, T$}
    \STATE $\Delta x^{t-1} = \operatorname{update} (x^{t-1},f, \texttt{update\_method\_name})$
    \STATE $\operatorname{L} = \texttt{loss\_function} (x^{t-1}, f)$ 
    % \STATE $\texttt{calculated by Eq.~\ref{eq:loss}}$
    \STATE $A^{t} = A^{t-1} + \Delta x^{t-1} \cdot \texttt{get\_grad}(x^{t-1},L)$
    \STATE $x^{t} = x^{t-1} + \Delta x^{t-1}$
\ENDFOR
\RETURN $A^T$
\end{algorithmic}
\end{algorithm}

\paragraph{Proof of Fundamental Attribution Method Satisfying Sensitivity Axiom and Complete Property}
\label{apx:pr-sens}

\begin{proof}
Let $x^0 = x'$ be the baseline input and $x^T = x$ be the target input. Suppose the path from $x'$ to $x$ is divided into $T$ segments, with $\{x^1, x^2, \ldots, x^{T-1}\}$ being intermediate points. Then, for each $t \in \{1, \ldots, T\}$:
\begin{equation}
\begin{aligned}
f\left(x^t\right) 
& = f\left(x^{t-1}\right) 
   + \frac{\partial f\left(x^{t-1}\right)}{\partial x^{t-1}}
   \bigl(x^t - x^{t-1}\bigr).
\end{aligned}
\end{equation}
Summing over $t$ from $1$ to $T$ gives:
\begin{equation}
\begin{aligned}
\sum_{t=1}^{T} f\left(x^t\right)
& = \sum_{t=1}^{T} \Bigl[f\left(x^{t-1}\right) + 
    \frac{\partial f\left(x^{t-1}\right)}{\partial x^{t-1}}
    \bigl(x^t - x^{t-1}\bigr)\Bigr] \\
& = \sum_{t=1}^{T} f\left(x^{t-1}\right)
    + \sum_{t=1}^{T} \frac{\partial f\left(x^{t-1}\right)}{\partial x^{t-1}}
    \bigl(x^t - x^{t-1}\bigr).
\end{aligned}
\end{equation}
Notice that $\sum_{t=1}^{T} f\bigl(x^{t-1}\bigr)$ telescopes with $\sum_{t=1}^T f\bigl(x^t\bigr)$ except for boundary terms:
\begin{equation}
\sum_{t=1}^{T} f\left(x^t\right) - \sum_{t=1}^{T} f\left(x^{t-1}\right)
= f\left(x^T\right) - f\left(x^0\right).
\end{equation}
Hence,
\begin{equation}
\begin{aligned}
f(x^T) - f(x^0)
& = \sum_{t=1}^{T} \frac{\partial f\left(x^{t-1}\right)}{\partial x^{t-1}}
    \bigl(x^t - x^{t-1}\bigr) \\
& = \sum_{t=1}^{T} \sum_{j=1}^d 
    \frac{\partial f\left(x^{t-1}\right)}{\partial x_j^{t-1}}
    \bigl(x_j^t - x_j^{t-1}\bigr) \\
& = \sum_{j=1}^d A_j\bigl(x, x'\bigr),
\end{aligned}
\end{equation}
where $A_j(x, x')$ denotes the total attribution for feature $j$ along the path from $x=x^0$ to $x'=x^T$.

Since the sum of the feature attributions 
$\sum_{j=1}^d A_j(x, x')$ equals $f(x) - f(x')$, the method inherently satisfies the Complete property. Moreover, if $f(x) \neq f(x')$, at least one feature $j$ must have a non-zero attribution $A_j(x, x')$, thereby satisfying the Sensitivity axiom Complete property.
\end{proof}

\paragraph{Satisfaction of the Implementation Invariance Axiom} As defined earlier in Section~\ref{sec:pd}, the Implementation Invariance axiom ensures that if two models produce identical outputs for all inputs, their attribution results should also be the same. Therefore, by accumulating the gradients at each step, the final attribution results $A(x, x')_{f_1}$ and $A(x, x')_{f_2}$ will remain consistent, ensuring that the attribution process is unaffected by the internal implementation details of the model.

\begin{table*}[t]
\caption{Explanation of Interpretability Methods with Advantages and Disadvantages}
\label{tab:explain-model}
\resizebox{\linewidth}{!}{%
\begin{threeparttable}
\begin{tabular}{@{}l|c|c|c|l|l|l@{}}
\toprule
Method & Sens & Impl & Comp & Principle & Advantages & Disadvantages \\ \midrule
DeepLIFT~\cite{shrikumar2017learning} & \pie{360} & \pie{0} & \pie{0} & \parbox{7cm}{By comparing the activation differences between the input and a reference baseline, the contribution values are propagated layer by layer through the neural network, calculating the relative importance of each input feature for the output.} & \parbox{6cm}{By comparing the activation differences between the input and baseline, it effectively avoids the vanishing gradient problem.} & \parbox{6cm}{Sensitive to baseline selection, which may lead to inconsistent attribution results.                            } \\ \midrule
SG~\cite{smilkov2017smoothgrad} & \pie{0} & \pie{0} & \pie{0} & \parbox{7cm}{By adding multiple random noises to the input to generate perturbed samples, the gradients of these samples are calculated and averaged, thereby smoothing the gradient-based attribution results.       } & \parbox{6cm}{Smoothing the gradient attribution results provides relatively intuitive explanations.} & \parbox{6cm}{Results depend on random noise, which may not fully satisfy attribution axioms, leading to inaccurate or inconsistent explanations.              } \\ \midrule
RISE~\cite{Petsiuk2018rise} & \pie{0} & \pie{0} & \pie{0} & \parbox{7cm}{By generating random masks on the input data and observing the changes in the model output, it calculates the importance score of each input feature to the model’s prediction, generating a heatmap.} & \parbox{6cm}{Does not require access to the internal model structure, suitable for any black-box model.} & \parbox{6cm}{High computational cost, relies on a large number of random mask samples, and struggles to capture complex interactions between input features.           } \\ \midrule
SM~\cite{simonyan2013deep} & \pie{0} & \pie{360} & \pie{0} & \parbox{7cm}{By calculating the gradient of the model output with respect to input features, it identifies the most important input areas for the model prediction.                 } & \parbox{6cm}{Simple and intuitive, low computational cost.} & \parbox{6cm}{Susceptible to gradient vanishing and noise, leading to unstable results and lack of precision.                 } \\ \midrule
Grad-CAM~\cite{selvaraju2017grad} & \pie{0} & \pie{0} & \pie{0} & \parbox{7cm}{By calculating the gradient of the target class with respect to the feature maps of convolutional layers, and weighting the feature maps to generate a heatmap, it highlights the most important input areas for the model prediction.   } & \parbox{6cm}{Generates intuitive visualization results, suitable for image tasks in convolutional neural networks.} & \parbox{6cm}{Does not satisfy attribution axioms, and lacks the ability to explain fine-grained features.                         } \\ \midrule
IG~\cite{sundararajan2017axiomatic} & \pie{360} & \pie{360} & \pie{360} & \parbox{7cm}{By calculating the gradient accumulation along a path between the input and reference baseline, it quantifies the contribution of each input feature to the model output.          } & \parbox{6cm}{Satisfies attribution axioms (Sensitivity and Implementation Invariance), providing consistent and reliable attribution results.} & \parbox{6cm}{Using linear zero vectors or black images as baselines leads to instability in attribution results.            } \\ \midrule
FIG~\cite{hesse2021fast} & \pie{360} & \pie{360} & \pie{360} & \parbox{7cm}{By reducing the number of path integration points, it provides a fast approximation of Integrated Gradients.            } & \parbox{6cm}{Improves computational efficiency while retaining a certain degree of attribution accuracy.} & \parbox{6cm}{Reducing the number of path integration points decreases the precision of results, which may lead to significant deviations in models with strong nonlinearity.      } \\ \midrule
GIG~\cite{kapishnikov2021guided} & \pie{360} & \pie{360} & \pie{360} & \parbox{7cm}{Combines IG and Guided Backpropagation by suppressing the negative gradient influence, highlighting the positive gradient contribution to the model output.} & \parbox{6cm}{Generates clearer and more intuitive attribution results.} & \parbox{6cm}{Suppressing negative gradients may lead to the loss of information from negatively correlated features, and the attribution results may not fully reflect the global influence of input features.   } \\ \midrule
EG~\cite{erion2021improving} & \pie{360} & \pie{360} & \pie{360} & \parbox{7cm}{By sampling the distribution of input features and averaging the Integrated Gradients results over multiple baselines, it computes a more stable attribution value.} & \parbox{6cm}{Reduces the impact of a single baseline choice on the attribution result, enhancing robustness and generalization ability to some extent.} & \parbox{6cm}{High computational cost, and additional information brought by sampling reduces the transparency of the explanation process.                 } \\ \midrule
BIG~\cite{wang2021robust} & \pie{360} & \pie{360} & \pie{360} & \parbox{7cm}{By dynamically selecting reference baselines and combining multi-path integration, it calculates the contribution of input features to the model output.                  } & \parbox{6cm}{Reduces dependency on a single baseline, improving the stability of attribution results to some extent.} & \parbox{6cm}{Dynamic baseline selection increases the computational complexity and uncertainty of the explanation process.                     } \\ \midrule
MFABA~\cite{zhu2024mfaba} & \pie{360} & \pie{360} & \pie{360} & \parbox{7cm}{By combining boundary point computation and adversarial perturbation exploration, it improves the accuracy and robustness of feature attribution.                   } & \parbox{6cm}{Increases the accuracy of attribution, enhancing the model's stability and robustness when facing adversarial samples.} & \parbox{6cm}{Relies on the basic boundary iteration method (BIM), limiting the diversity and range of path exploration, which may not fully capture feature changes under different input perturbations.} \\ \midrule
AGI~\cite{pan2021explaining} & \pie{360} & \pie{360} & \pie{0} & \parbox{7cm}{By constructing the steepest ascent path from target adversarial samples to the original input, combined with gradient information, it analyzes the decision change of the model with respect to input features, generating attribution results. } & \parbox{6cm}{Bypasses the dependence on a fixed baseline, revealing sensitive features of the model under adversarial perturbations through nonlinear path exploration, providing relatively robust and consistent explanations.} & \parbox{6cm}{Path selection may be limited to a single model's decision boundary, lacking cross-model consistency, and computational cost is high for complex inputs.   } \\ \midrule
AttEXplore~\cite{zhuattexplore} & \pie{360} & \pie{360} & \pie{360} & \parbox{7cm}{By exploring cross-model consistent paths in the multi-model parameter space, focusing on the role of adversarial attack transferability in attribution. } & \parbox{6cm}{Provides more robust and comprehensive explanations, maintaining consistency across models and tasks, while enhancing resistance to noise and perturbations through frequency domain analysis.} & \parbox{6cm}{High computational complexity, path exploration requires significant computational resources.                         } \\ \midrule
LA~\cite{zhu2024enhancing} & \pie{360} & \pie{360} & \pie{0} & \parbox{7cm}{By analyzing the impact of local perturbations on input features, it quantifies the importance of each feature under specific input conditions.          } & \parbox{6cm}{Provides more granular explanations, efficient computation, and accurately reveals the model’s dependence on specific input features.} & \parbox{6cm}{Lacks a comprehensive capture of global feature interactions, potentially overlooking the synergistic effects between features.             } \\ \midrule
ISA~\cite{zhuiterative} & \pie{360} & \pie{360} & \pie{0} & \parbox{7cm}{By iteratively applying gradient ascent and descent, it gradually extracts the importance of input features.                       } & \parbox{6cm}{By gradually discovering the most influential features, it ensures that later iterations reveal more important features than earlier ones, greatly improving attribution accuracy and providing more precise and comprehensive explanations.} & \parbox{6cm}{Iterative process has high computational overhead.                                    } \\ \midrule
Chefer et al.~\cite{chefer2021generic} & \pie{360} & \pie{360} & \pie{0} & \parbox{7cm}{By combining attention weights and gradients in Transformer models, it computes feature contributions and generates unified attribution results for multimodal and encoder-decoder tasks.} & \parbox{6cm}{Suitable for various Transformer architectures.} & \parbox{6cm}{Relies on attention weights, which may fail to fully capture the impact of non-attention mechanisms on model decisions.              } \\ \midrule
M2IB~\cite{wang2023visual} & \pie{0} & \pie{0} & \pie{0} & \parbox{7cm}{By maximizing the correlation between input modality features and model outputs in the Information Bottleneck framework, it captures information that maximally influences the prediction, filtering out irrelevant information.} & \parbox{6cm}{Provides clear identification of important input features and ensures that only relevant features are retained.} & \parbox{6cm}{Requires large-scale training data, which limits its application on small-scale datasets or under noisy environments.                    } \\ \midrule

\end{tabular}%
\begin{tablenotes}
\item \textbf{\pie{360}}: The method satisfies the attribution axiom; \textbf{\pie{0}}: The method does not satisfy the attribution axiom.
\end{tablenotes}
\end{threeparttable}
}
\end{table*}

\paragraph{Update Methods} Update methods can be broadly defined as any mechanism that modifies the sample. However, to ensure consistency with the attribution method, certain restrictions must be applied to the update methods. Specifically, when a sample undergoes a small change that results in a significant impact on the model's output, it indicates that the features responsible for these changes play an important role in the model's decision. Therefore, theoretically, any update method can satisfy the \textbf{attribution axioms}, but here we introduce the \textbf{adversarial attack method}, as it has been shown to be particularly effective at minimizing sample changes while maximizing changes in the model's output. Moreover, the use of adversarial attack updates guarantees that interpretability methods will will avoid the issues of \textbf{impossibility theorems}, as these methods inherently do not satisfy the Linear. This failure to meet the Linear directly contributes to the failure to satisfy other key impossibility theorems, which demonstrates that attribution methods will perform better than random selection~\cite{bilodeau2024impossibility}.

As shown in Equation \ref{eq:attack}, the adversarial attack method, by applying minimal perturbations to the input sample, maximizes the change in the model's output. Equation \ref{eq:attack} represents the difference between the model outputs $f(x^{adv})$ and $f(x)$ after applying perturbation $\epsilon$ to the input sample $x$, where the direction of the perturbation is determined by the gradient's sign. Specifically, the direction of the perturbation is determined by the sign of $\nabla_x f(x)$, which is opposite to the gradient direction, thereby producing the maximum influence on the model's decision within the smallest perturbation range. Therefore, adversarial attacks are theoretically the smallest perturbations that have the greatest impact on the model's predictions.

\begin{equation}
f(x^{adv}) - f(x) = \epsilon \cdot \nabla_x f(x) \cdot \text{sign}(\nabla_x f(x))
\label{eq:attack}
\end{equation}

In the specific application of adversarial attacks, there are two main methods. The first is to accumulate gradients along the linear path between the original sample and the adversarial sample after the attack is completed \cite{wang2021robust}. The second method is to accumulate gradients while executing the attack \cite{zhuattexplore}, which uses a nonlinear exploration path. In contrast, methods that do not use adversarial attacks for updates often rely on baseline samples that are independent of the sample's information, such as using an all-black sample as a reference \cite{sundararajan2017axiomatic}. Alternatively, custom update methods can also be employed, although adversarial attacks generally offer a more efficient means of achieving this objective.

\paragraph{Justification for Using \texorpdfstring{$\text{sign}(\nabla_x f(x))$}{sign()}}

We now provide a theoretical justification for the use of the $\text{sign}(\nabla_x f(x))$ function in attribution and update procedures. The goal is to determine a perturbation direction $\Delta x$ that maximizes the model output change while ensuring fairness—i.e., each feature is perturbed with equal magnitude~\cite{zhu2024iterative}.

Assume a differentiable model $f: \mathbb{R}^n \to \mathbb{R}$ and an input $x \in \mathbb{R}^n$. Let $\Delta x$ be a small perturbation to $x$. According to the first-order Taylor expansion, we have:
\begin{equation}
f(x + \Delta x) - f(x) \approx \nabla_x f(x)^\top \Delta x.
\end{equation}

To ensure fairness and bounded perturbations, we impose an $\ell_\infty$ constraint:
\begin{equation}
\|\Delta x\|_\infty \leq \epsilon,
\end{equation}
which restricts each individual feature perturbation to lie in the range $[-\epsilon, \epsilon]$.

\textbf{Objective:} Maximize the output change under the $\ell_\infty$ constraint:
\begin{equation}
\max_{\|\Delta x\|_\infty \leq \epsilon} \nabla_x f(x)^\top \Delta x = \sum_{i=1}^n \frac{\partial f(x)}{\partial x_i} \cdot \Delta x_i.
\end{equation}

This is a convex optimization problem with a known closed-form solution. For each component $i$, since all feature perturbations are bounded by the same maximum magnitude due to the $\ell_\infty$ constraint, the optimal solution simply takes the largest allowable step in the direction of the gradient. That is, the optimal perturbation is given by:

\begin{equation}
\Delta x_i^* = \epsilon \cdot \text{sign}\left( \frac{\partial f(x)}{\partial x_i} \right),
\end{equation}
which leads to the optimal perturbation vector:
\begin{equation}
\Delta x^* = \epsilon \cdot \text{sign}(\nabla_x f(x)).
\end{equation}

Substituting into the objective, we obtain the maximum possible change in output:
\begin{equation}
\max_{\|\Delta x\|_\infty \leq \epsilon} \nabla_x f(x)^\top \Delta x = \epsilon \cdot \|\nabla_x f(x)\|_1.
\end{equation}

\textbf{Conclusion:} The use of $\text{sign}(\nabla_x f(x))$ ensures that each feature receives an update of identical magnitude, promoting fairness, while also maximizing the model’s output change under a fixed perturbation budget. This result theoretically validates the choice of $\text{sign}(\cdot)$ as the most effective direction for attribution and adversarial update methods under $\ell_\infty$ constraints.

\paragraph{Robustness Module} Because adversarial attacks are more effective in updating samples, we provide a robustness module. The robustness module covers both white-box and black-box attack methods. White-box attacks consider the parameters of a single model, which may lead to overfitting and limit the model's generalization ability. In contrast, black-box attacks account for parameter differences between different models and adapt to different decision boundaries, offering better generalization \cite{zhuattexplore}. Both white-box and black-box attacks can serve as update methods, with black-box attacks often performing better in terms of model interpretability, as shown in the experimental results in Section~\ref{sec:transfer}. Additionally, this module can independently evaluate a model's robustness, allowing researchers to test and compare the robustness of different models as needed.

\paragraph{Interpretability Module} This module specifically focuses on compliance with attribution axioms to ensure that the model's explanations are consistent and reliable. By comparing the advantages and disadvantages of different methods, researchers can select the most suitable interpretability method based on their needs, thereby improving the understanding of the model's decision-making process. Furthermore, the extension of this module to a variety of tasks and models increases its applicability, making it suitable for model interpretation and analysis across different domains.

Table~\ref{tab:explain-model} lists the interpretability methods we have implemented, including many state-of-the-art methods that satisfy the attribution axioms, along with several classical interpretability methods that do not. The table provides details on whether each method satisfies the Sensitivity (Sens) and Implementation Invariance (Impl) axioms, outlines the algorithmic principles behind each method, and we also summarize their advantages and disadvantages in Table~\ref{tab:explain-model}. In addition, we also consider the Complete property (Comp) in our evaluation, which refers to an attribution method’s ability to accurately assign the entire output of the model to its input features. We use the notation \textbf{\pie{360}} to denote that a method satisfies the corresponding attribution axiom, and \textbf{\pie{0}} to denote that it does not. The Complete property ensures that the sum of the attribution values for all input features reconstructs the model's output. While the Complete property is not as critical as Sensitivity, it serves as a useful reference for evaluating attribution methods. Notably, any method satisfying the Complete property must also satisfy Sensitivity; however, satisfying Sensitivity does not necessarily imply that a method satisfies the Complete property. Some methods, such as AGI, LA, and ISA, only satisfy the Complete property locally, rather than globally. To ensure completeness of our framework, we have extended these methods to applications in image classification, text classification, multimodal tasks, and object detection tasks. The related experimental results, which highlight the performance and limitations of each method.

\paragraph{Validation Module}

In this study, we adopted several evaluation metrics to quantify the effectiveness of the interpretability methods and selected appropriate metrics based on the characteristics of the tasks.

For the image classification task, we primarily used Insertion Score (INS), Deletion Score (DEL), INFD Score (INFD), Frames Per Second (FPS), and Instances Per Second (IPS). Insertion Score measures the change in model decision when a feature is added, reflecting the importance of that feature for the decision. Deletion Score measures the change in model decision when a feature is removed, indicating the model's dependency on that feature. INFD Score quantifies the change in model uncertainty after a feature is added or removed, with higher values indicating greater uncertainty and weaker interpretability. Frames Per Second measures the number of explanations generated per second, indicating the speed of the model in generating explanations, suitable for real-time applications.

For the text classification task, in addition to Insertion Score and Deletion Score, we also introduced Instances Per Second (IPS) to measure the number of text instances processed per second, reflecting the efficiency of text processing and explanation.

\begin{lstlisting} [caption=Example Code for Image Classification Task, basicstyle=\footnotesize\ttfamily]
from demo.algorithm.explanation import *
# First, define the explanation task
from demo.task import ExplanationTask

# The explanation task requires the loss function and forward function
def loss_fn(batch):
   x, y = batch
   logits = model(x)
   loss = -torch.diag(logits[:, y]).sum()
   return loss

def forward(batch):
   x, _ = batch
   return model(x)

# Define the explanation task
explanation_task = ExplanationTask(loss_fn=loss_fn, forward_fn=forward, model_type=ModelType.IMAGECLASSIFICATION)

# Then, define the explanation algorithm
explanation = IG(explanation_task)

# Obtain the attribution result
attribution = explanation([sample_x, sample_y])
\end{lstlisting}

For the multimodal task, we introduced several specially designed metrics: Image Confidence Drop (ICD) and Image Confidence Increase (ICI), which measure the decrease and increase in model confidence for image-related information during the explanation process, respectively. Similarly, Text Confidence Drop (TCD) and Text Confidence Increase (TCI) measure the confidence changes in text-related information during the explanation process. Higher values indicate weaker or stronger interpretability for the respective information. These metrics help us better understand the model's ability to explain images and text in multimodal tasks. Additionally, Frames Per Second is used as a general metric to measure the explanation generation speed for multimodal tasks.

For the object detection task, the evaluation metrics include Insertion Score (INS), Deletion Score (DEL), and Frames Per Second (FPS). These metrics help us measure the importance, dependence, and real-time explanation generation of features for object detection models.

Through these evaluation metrics, we can comprehensively assess the interpretability performance of different tasks and models, ensuring the effectiveness and applicability of the proposed method.

\begin{table*}[t]
\centering
\caption{Evaluation Metrics Applied to Different Tasks (Image, Text, Multimodal(MM), and Object Detection(OD)) and Their Descriptions}
\label{tab:matrics}
\resizebox{\textwidth}{!}{%
\begin{threeparttable}
\begin{tabular}{@{}l|c|c|c|c|c|l@{}}
\toprule
Metric                      & Abbr.   & Image               & Text               & MM                  & OD                 & Metric Description                                                   \\ \midrule
Insertion Score           & INS  & \pie{360} & \pie{360} & \pie{0}   & \pie{360} & \parbox{10cm}{Measures the change in model decision after adding certain features, with higher values indicating that the interpretability method better captures the importance of the feature.}             \\ \midrule
Deletion Score            & DEL  & \pie{360} & \pie{360} & \pie{0}   & \pie{360} & \parbox{10cm}{Measures the change in model decision after removing certain features, with lower values indicating that the interpretability method better removes important features.}              \\ \midrule
INFD Score                & INFD & \pie{360} & \pie{0}   & \pie{0}   & \pie{0}   & \parbox{10cm}{Demonstrates the highest faithfulness based on these tests.}            \\ \midrule
Image Confidence Drop     & ICD  & \pie{0}   & \pie{0}   & \pie{360} & \pie{0}   & \parbox{10cm}{Measures the decline in model confidence for image-related information during explanation, with lower values indicating better removal of perturbation features and stronger interpretability.} \\ \midrule
Image Confidence Increase & ICI  & \pie{0}   & \pie{0}   & \pie{360} & \pie{0}   & \parbox{10cm}{Measures the increase in model confidence for image-related information during explanation, with higher values indicating that the interpretability method better highlights important features.} \\ \midrule
Text Confidence Drop      & TCD  & \pie{0}   & \pie{0}   & \pie{360} & \pie{0}   & \parbox{10cm}{Measures the decline in model confidence for text-related information during explanation, with lower values indicating better removal of perturbation features and stronger interpretability.} \\ \midrule
Text Confidence Increase  & TCI  & \pie{0}   & \pie{0}   & \pie{360} & \pie{0}   & \parbox{10cm}{Measures the increase in model confidence for text-related information during explanation, with higher values indicating that the interpretability method better highlights important features.} \\ \midrule
Frames Per Second         & FPS  & \pie{360} & \pie{0}   & \pie{360} & \pie{360} & \parbox{10cm}{Measures the number of explanations generated per second by the model, with higher values indicating faster explanation generation, suitable for real-time applications.}             \\ \midrule
Instances Per Second      & IPS  & \pie{0}   & \pie{360} & \pie{0}   & \pie{0}   & \parbox{10cm}{Measures the number of text instances that can be processed per second, with higher values indicating faster text processing and explanation generation.}                 \\ \bottomrule
\end{tabular}%

\begin{tablenotes}
\item \textbf{\pie{360}}: The metric is used in this task; \textbf{\pie{0}}: The metric is not used in this task
\end{tablenotes}

\end{threeparttable}
}
\end{table*}

\paragraph{Data \& Model Module} To facilitate user access, we provide data \& model loading modules for each task, along with example files in the code repository for reference. In this Demo 1, the \lstinline|ExplanationTask| class takes the loss function and forward function as initialization parameters to ensure that the task-specific loss and predictions can be calculated. Next, we define an explanation algorithm (e.g., \lstinline|IG|) and generate attribution results by providing sample inputs. For other task types (e.g., text classification, multimodal tasks, and object detection), users can refer to additional examples in the anonymous repository.

\section{Experiments}

\subsection{Models and Datasets}\label{apx:exp-settings}
In this experiment, we selected several models and datasets to evaluate the performance of the proposed method. For the image classification task, we used five commonly used Deep Neural Network (DNN) architectures: Inception-v3~\cite{szegedy2016rethinking}, ResNet-50~\cite{he2016deep}, VGG16~\cite{simonyan2014very}, MaxViT-T, and ViT-B/16\cite{dosovitskiy2020image}. These models have different architectures and depths, covering traditional deep CNNs and more recent vision Transformer architectures. The ImageNet~\cite{ILSVRC15} dataset was used for the image classification task, as it contains a large number of images suitable for large-scale visual recognition tasks.

For the text classification task, we used two models: TextCNN~\cite{zhang2015sensitivity} and GRU~\cite{chung2014empirical}. TextCNN is a convolutional neural network-based model for text classification that can effectively capture local features in text, while GRU is a common recurrent neural network structure suitable for processing sequential data. The IMDb and Yelp Reviews datasets were used for the text classification task, both widely used for sentiment analysis and text classification tasks.

For the multimodal task, we selected the CLIP model~\cite{radford2021learning}, which can handle both image and text data, making it suitable for multimodal learning tasks. We validated the model on several datasets, including Conceptual Captions~\cite{sharma2018conceptual}, ImageNet~\cite{ILSVRC15}, and Flickr8k~\cite{hodosh2013framing}. These datasets contain rich image-text pairs suitable for tasks such as image captioning and image-text matching.

Finally, for the object detection task, we used the RetinaNet~\cite{lin2017focal} model, which performs well in object detection and can effectively identify and locate objects within images. The COCO dataset~\cite{lin2014microsoft}, which contains various types of images and objects, was used for the object detection task. It is one of the standard datasets in the field of object detection.

As shown in Table~\ref{tab:exp-setting}, the models and datasets used in this experiment are listed in detail and categorized according to task type.

\begin{table}[t]
\centering
\caption{Models and datasets used across different experimental scenarios.}
\label{tab:exp-setting}
\resizebox{\linewidth}{!}{%
\begin{tabular}{@{}p{2.5cm}|p{2.5cm}|p{3.5cm}@{}}
\toprule
\textbf{Task} & \textbf{Datasets} & \textbf{Models} \\ \midrule
\textbf{Image Classification} 
& ImageNet 
& Inception-v3, ResNet-50, VGG16, MaxViT-T, ViT-B/16 \\ \midrule
\textbf{Text Classification} 
& IMDb, Yelp Reviews 
& TextCNN, GRU \\ \midrule
\textbf{Multimodal Tasks} 
& Conceptual Captions, ImageNet, Flickr8k 
& CLIP (ViT-B/32) \\ \midrule
\textbf{Object Detection} 
& COCO 
& RetinaNet \\
\bottomrule
\end{tabular}
}
\end{table}

\subsection{Interpretability Module}

To evaluate the effectiveness of the proposed framework, we compared it with several existing interpretability methods as baselines in our experiments. These methods include FIG~\cite{hesse2021fast}, DeepLIFT~\cite{shrikumar2017learning}, GIG~\cite{kapishnikov2021guided}, IG~\cite{sundararajan2017axiomatic}, EG~\cite{erion2021improving}, SG~\cite{smilkov2017smoothgrad}, BIG~\cite{wang2021robust}, SM~\cite{simonyan2013deep}, MFABA~\cite{zhu2024mfaba}, AGI~\cite{pan2021explaining}, AttEXplore~\cite{zhuattexplore}, LA~\cite{zhu2024enhancing}, ISA~\cite{zhuiterative}, Chefer et al.~\cite{chefer2021generic}, and Grad-CAM~\cite{selvaraju2017grad}. By applying these methods to various deep learning models and tasks, including image classification, object detection, and text classification, we assess their performance, demonstrating the advantages and effectiveness of our framework in multiple scenarios.

\begin{table*}[t]
\caption{Experimental Results for Image Classification Task}
\label{tab:img-cls-result}
\resizebox{\textwidth}{!}{%

\begin{threeparttable}

\begin{tabular}{@{}c|c|cccc|cccc|cccc|cccc|cccc@{}}
\toprule
                                                             &                            & \multicolumn{4}{c|}{Inception-v3}                       & \multicolumn{4}{c|}{ResNet-50}                           & \multicolumn{4}{c|}{VGG16}                               & \multicolumn{4}{c|}{MaxViT-T}                           & \multicolumn{4}{c}{ViT-B/16}                            \\ \midrule
Method                                                       & Axioms                     & INS & DEL & INFD & FPS     & INS & DEL & INFD & FPS      & INS & DEL & INFD & FPS      & INS & DEL & INFD & FPS     & INS & DEL & INFD & FPS     \\\midrule
DeepLIFT                                                     & \pie{0}   & 0.295& 0.042& 169.15& 3.07& 0.125& 0.031& 154.46& 23.24& 0.093& 0.023& 292.33& 20.88& 0.499& 0.182& 7.28& 11.10& 0.322& 0.084& 10.03& 27.21
\\
SG                                                           & \pie{0}   & 0.389& \textbf{0.033}& 105.23& 10.40& 0.277& \textbf{0.023}& 54.88& 12.13& 0.186& \textbf{0.016}& 97.40& 6.52& 0.642& \textbf{0.140}& 3.53& 2.38& 0.416& \textbf{0.048}& 3.65& 2.90
\\
SM                                                           & \pie{0}   & 0.533& 0.063& 176.11& 14.05& 0.315& 0.057& 169.15& 24.76& 0.270& 0.042& 311.83& 19.56& 0.490& 0.196& 7.95& 5.33& 0.337& 0.098& 10.30& 32.61
\\
Chefer et al. & \pie{360}   & 0.295& 0.044& 33.58& 24.45& 0.218& 0.031& 5.75& 44.76& 0.183& 0.028& 26.01& 50.71& 0.472& 0.182& 0.57& 10.96& 0.318& 0.087& 0.92& 28.00
\\
Grad-CAM                                                     & \pie{0}   & 0.576& 0.166& 21.79& \textbf{56.95}& \textbf{0.614}& 0.163& 4.82& \textbf{143.79}& 0.498& 0.135& 7.78& \textbf{129.31}& \textbf{0.744}& 0.332& 0.56& \textbf{33.50}& 0.282& 0.296& 1.53& \textbf{74.32}
\\
FIG                                                          & \pie{360} & 0.202& 0.046& 173.04& 24.43& 0.106& 0.032& 161.32& 46.06& 0.079& 0.027& 301.12& 39.38& 0.462& 0.182& 7.32& 11.67& 0.268& 0.088& 10.02& 25.38
\\
GIG                                                          & \pie{360} & 0.319& 0.034& 87.24& 0.77& 0.145& 0.019& 61.28& 1.29& 0.103& 0.017& 94.91& 1.48& 0.545& 0.139& 3.07& 0.64& 0.359& 0.064& 4.14& 1.14
\\
IG                                                           & \pie{360} & 0.321& 0.043& 88.76& 6.83& 0.145& 0.028& 87.76& 11.86& 0.096& 0.023& 139.71& 6.46& 0.542& 0.188& 8.13& 1.96& 0.378& 0.068& 4.12& 2.88
\\
EG                                                           & \pie{360} & 0.376& 0.265& 174.73& 21.93& 0.350& 0.283& 174.49& 58.31& 0.357& 0.337& 314.70& 41.87& 0.599& 0.515& 8.56& 19.81& 0.434& 0.382& 9.82& 48.50
\\
BIG                                                          & \pie{360} & 0.484& 0.054& 13.12& 0.07& 0.291& 0.047& 1.79& 0.13& 0.227& 0.037& 17.34& 0.15& 0.568& 0.187& 0.59& 0.03& 0.471& 0.113& 0.95& 0.09
\\
MFABA                                                        & \pie{360} & 0.539& 0.064& 20.78& 5.23& 0.320& 0.056& 2.86& 8.83& 0.279& 0.041& 15.59& 5.42& 0.441& 0.358& 0.57& 3.97& 0.398& 0.184& 0.92& 8.09
\\
AGI                                                          & \pie{360} & 0.572& 0.058& 5.63& 0.04& 0.501& 0.051& 1.03& 0.06& 0.397& 0.042& 0.98& 0.03& 0.645& 0.198& 0.57& 0.02& 0.500& 0.085& 0.92& 0.04
\\
AttEXplore                                                   & \pie{360} & 0.619& 0.044& 4.83& 0.07& 0.504& 0.033& 0.86& 0.12& 0.443& 0.028& 0.61& 0.07& 0.616& 0.158& \textbf{0.56}& 0.05& 0.524& 0.077& \textbf{0.91}& 0.09
\\
LA                                                           & \pie{360} & 0.646& 0.068& 5.39& 0.02& 0.550& 0.047& 0.88& 0.04& 0.437& 0.034& 1.01& 0.02& 0.705& 0.209& \textbf{0.56}& 0.01& 0.607& 0.073& 0.96& 0.02
\\
ISA                                                          & \pie{360} & \textbf{0.723}& 0.067& \textbf{4.82}& 0.03& 0.604& 0.051& \textbf{0.83}& 0.04& \textbf{0.509}& 0.042& \textbf{0.56}& 0.03& 0.738& 0.250& \textbf{0.56}& 0.02& \textbf{0.658}& 0.124& \textbf{0.91}& 0.03\\ \bottomrule
\end{tabular}%
\begin{tablenotes}
\item \textbf{\pie{360}}: The method satisfies the attribution axioms; \textbf{\pie{0}}: The method does not satisfy the attribution axioms.
\end{tablenotes}
\end{threeparttable}

}
\end{table*}

\begin{table}[t]
\centering
\caption{Experimental Results for Text Classification Task}
\label{tab:txt-cls-result}
\resizebox{0.9\linewidth}{!}{%
\begin{tabular}{@{}c|c|ccc|ccc@{}}
\toprule
\multirow{2}{*}{Model}   & \multirow{2}{*}{Method}                                      & \multicolumn{3}{c|}{IMDb}  & \multicolumn{3}{c}{Yelp Reviews} \\ \cmidrule(l){3-8} 
                         &                                                              & INS    & DEL    & IPS      & INS      & DEL      & IPS        \\ \midrule
\multirow{9}{*}{GRU}     & IG                                                           & 0.907& \textbf{0.095}& 0.60& 0.733& \textbf{0.107}& 4.22
\\
                         & SM                                                           & 0.818& 0.701& 148.16& 0.801& 0.419& 182.01
\\
                         & SG                                                           & 0.829& 0.807& 4.31& 0.687& 0.677& 4.31
\\
                         & LA                                                           & 0.841& 0.823& 7.07& 0.771& 0.693& 7.10
\\
                         & MFABA                                                        & 0.881& 0.519& 5.25& 0.810& 0.361& 5.25
\\
                         & AGI                                                          & 0.898& 0.598& 5.28& 0.824& 0.463& 5.25
\\
                         & FIG                                                          & 0.907& 0.095& \textbf{218.32}& 0.720& 0.125& 194.59
\\
                         & AttEXplore                                                   & 0.910& 0.478& 0.89& 0.819& 0.363& 0.89
\\
                         & Chefer et al. & \textbf{0.922}& 0.206& 179.69& \textbf{0.807}& 0.237& \textbf{206.97}
\\ \midrule
\multirow{9}{*}{TextCNN} & IG                                                           & 0.902& 0.102& 1.54& 0.756& 0.122& 13.12
\\
                         & SM                                                           & \textbf{0.903}& 0.608& 254.83& \textbf{0.810}& 0.388& 470.64
\\
                         & SG                                                           & 0.803& 0.836& 4.92& 0.681& 0.676& 13.33
\\
                         & LA                                                           & 0.802& 0.791& 20.60& 0.598& 0.693& 11.50
\\
                         & MFABA                                                        & 0.894& 0.573& 12.84& 0.803& 0.388& 7.80
\\
                         & AGI                                                          & 0.667& 0.902& 12.31& 0.464& 0.812& 16.70
\\
                         & FIG                                                          & 0.902& \textbf{0.100}& \textbf{612.22}& 0.755& \textbf{0.117}& \textbf{533.73}
\\
                         & AttEXplore                                                   & 0.896& 0.632& 0.48& 0.802& 0.434& 2.23
\\
                         & Chefer et al. & 0.901& 0.244& 312.84& 0.793& 0.223& 411.17\\ \bottomrule
\end{tabular}%
}
\end{table}

\paragraph{Image Classification Task} 
In the analysis of interpretability methods for image classification, methods adhering to attribution axioms generally offer better interpretability and consistency, while non-axiomatic methods tend to be more computationally efficient. As shown in Table~\ref{tab:img-cls-result}, axiom-compliant methods like ISA and AttEXplore achieve higher insertion scores and lower deletion scores, indicating stronger interpretability. ISA, in particular, excels by maintaining high insertion scores, low deletion scores, and a smaller INFD value, demonstrating its ability to capture key model information effectively.

On the other hand, non-axiomatic methods such as Grad-CAM and DeepLIFT are more efficient, with Grad-CAM delivering high FPS across most models. While their interpretability is weaker compared to axiom-compliant methods, their lower computational cost makes them suitable for resource-constrained environments.

As model complexity increases, axiom-compliant methods, especially ISA, outperform non-axiomatic ones in interpretability, particularly in transformer-based models. Thus, selecting an interpretability method requires balancing explanation accuracy, computational efficiency, and model complexity.

\begin{table}[t]
\centering
\caption{Experimental Results for Object Detection Task}
\label{tab:ob-result}
\resizebox{0.6\linewidth}{!}{%
\begin{tabular}{@{}c|c|c|c@{}}
\toprule
Method                                                       & INS    & DEL    & FPS    \\ \midrule
FIG                                                          & 0.5219 & 0.3236 & 0.2162 \\
IG                                                           & 0.5432 & 0.3747 & 0.0053 \\
SG                                                           & 0.5747 & 0.3705 & 0.0055 \\
Chefer et al. & 0.5530 & 0.2691 & 0.2564 \\
AttEXplore                                                   & 0.6508 & \textbf{0.2357} & 0.0018 \\
SM                                                           & \textbf{0.6324} & 0.2478 & \textbf{0.4025} \\
MFABA                                                        & 0.6249 & 0.2574 & 0.0105 \\ \bottomrule
\end{tabular}%
}
\end{table}

\begin{figure*}[t]
    \centering
    \includegraphics[width=\linewidth]{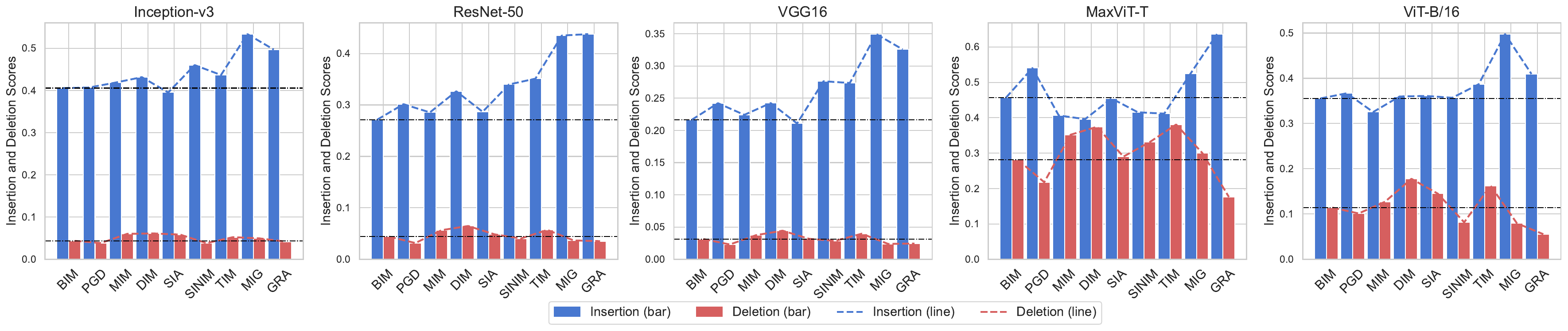}
    \caption{Impact of Different Update Methods on Attribution Performance}
    \label{fig.transfer}
\end{figure*}

\paragraph{Text Classification Tasks}
In text classification tasks, as shown in Table~\ref{tab:txt-cls-result}, IG, FIG, and Chefer et al.~\cite{chefer2021generic} demonstrate the best interpretability, especially in terms of high insertion scores and low deletion scores. In comparison, other methods that are not specifically optimized for text explanation tend to have weaker interpretability, characterized by higher deletion scores. In terms of computational efficiency, SM and FIG perform the best in terms of instances per second (IPS), indicating strong computational speed. However, considering both interpretability and computational efficiency, FIG performs the best in text classification tasks, achieving the highest insertion score, the lowest deletion score, and the fastest IPS, demonstrating its dual advantage in both explanation quality and efficiency.
\begin{table*}[h]
\centering
\caption{Experimental Results for Multimodal Task}
\label{tab:mm-result}
\resizebox{.8\linewidth}{!}{%
\begin{tabular}{@{}c|ccccc|ccccc|ccccc@{}}
\toprule
                                                             & \multicolumn{5}{c|}{Conceptual Captions}  & \multicolumn{5}{c|}{ImageNet}          & \multicolumn{5}{c}{Flickr8k}            \\ \midrule
Method                                                       & ICD     & ICI   & TCD    & TCI   & FPS    & ICD    & ICI  & TCD    & TCI  & FPS    & ICD     & ICI  & TCD    & TCI  & FPS    \\ \midrule
M2IB                                                         & \textbf{1.12}& \textbf{39.30}& 1.71& 37.40& 0.66& 1.16& 49.40& 2.60& 25.40& 0.80& \textbf{1.47}& \textbf{28.10}& 2.08& 34.70& 0.74
\\
RISE                                                         & 1.42& 28.80& \textbf{0.80}& 43.95& 0.10& \textbf{1.00}& \textbf{54.00}& \textbf{0.99}& \textbf{46.80}& 0.11& 3.01& 5.70& \textbf{0.89}& \textbf{46.40}& 0.11
\\
Chefer et al. & 2.01& 33.65& 0.93& \textbf{45.30}& \textbf{1.27}& 1.66& 44.00& 1.67& 29.90& \textbf{2.79}& 2.62& 26.80& 1.36& 42.60& \textbf{2.46}
\\
Grad-CAM                                                     & 4.11& 20.20& 1.80& 34.40& 1.17& 2.55& 33.90& 2.64& 25.70& 2.31& 5.19& 13.60& 2.18& 34.20& 1.96
\\
SM                                                           & 10.44& 2.95& 1.07& 40.05& 0.93& 4.73& 16.40& 1.76& 33.10& 1.57& 12.15& 0.10& 1.08& 45.90& 1.40
\\
FIG                                                          & 10.51& 2.90& 0.97& 41.25& 0.94& 4.79& 16.90& 1.65& 34.80& 1.54& 12.22& 0.10& 1.31& 43.90& 1.39\\ \bottomrule
\end{tabular}%
}
\end{table*}

\paragraph{Object Detection Tasks}

In object detection tasks, methods that follow attribution axioms continue to demonstrate superior interpretability. As shown in Table~\ref{tab:ob-result}, AttEXplore and MFABA exhibit good insertion and deletion scores, especially in deletion scores, where they manage to maintain low values. This suggests their ability to retain key information when removing irrelevant features, while also offering stable computational efficiency. Notably, AttEXplore, despite its slower computational speed, demonstrates excellent interpretability.

Additionally, the SM method also shows good interpretability and the fastest computational speed (FPS), making it highly practical for object detection tasks. SM strikes a good balance between insertion and deletion scores and offers the fastest computational efficiency compared to other methods, making it suitable for applications that require fast feedback while maintaining interpretability. 

\paragraph{Multimodal Tasks}
As shown in Table~\ref{tab:mm-result}, the M2IB~\cite{wang2023visual} method outperforms others in multimodal tasks, providing balanced performance in both image and text confidence changes, as well as strong insertion and deletion scores across datasets. Specifically, on Conceptual Captions and Flickr8k, M2IB shows significant confidence increases and minimal decreases, while maintaining stable computational efficiency.

RISE~\cite{Petsiuk2018rise} and Chefer et al.~\cite{chefer2021generic} offer reasonable interpretability but are not optimized for multimodal tasks. They perform well in confidence increase but lag behind M2IB in interpretability and computational efficiency. Methods like Grad-CAM, MFABA, SM, and FIG, which are not designed for multimodal tasks, show suboptimal interpretability, particularly in image and text confidence decrease metrics.

\subsection{Robustness Module}
\label{sec:transfer}

Additionally, we provide several robustness methods, including AdvGAN, BIM, C\&W, DIM, FGSM, MIM, NAA, PGD, SINIM, SSA, TIM, SIA, MIG, and GRA. We explored the effects of combining some of these methods with AttEXplore to examine how different robustness methods impact interpretability. As shown in Figure~\ref{fig.transfer}, we combined various robustness modules with the AttEXplore attribution method to explore updated samples. The results indicate significant changes in model interpretability when different robustness modules are used for exploration. Specifically, the two black dashed lines parallel to the x-axis in Figure~\ref{fig.transfer} represent the insertion and deletion scores when AttEXplore uses the basic BIM update method, which serves as the baseline performance. When using different robustness methods for updating, we observe that stronger robustness modules lead to corresponding improvements in AttEXplore's interpretability. Particularly, the MIG and GRA modules, which are relatively powerful, outperform the BIM baseline in terms of interpretability across all models.

\section{Conclusion}

In this work, we present \textbf{ABE}, a unified and extensible framework for Attribution-Based Explainability. ABE establishes a general theoretical foundation through \emph{Fundamental Attribution Methods}, enabling the integration of diverse update strategies while ensuring adherence to key attribution axioms. In addition, ABE incorporates robustness module that support faithful and stable attribution under adversarial perturbations, offering a principled approach to explanation evaluation. Its modular design supports a wide range of models and tasks, facilitating the development, comparison, and deployment of axiom-compliant and robustness-enhanced attribution methods. We envision ABE as a foundation for future research in interpretable machine learning, particularly in scenarios involving complex models, multimodal data, and real-world deployment where explanation consistency and reliability are critical.

\bibliography{main}
\bibliographystyle{ieeetr}

\end{document}